\documentclass[5p,times]{elsarticle}

\usepackage{graphicx}
\usepackage{algorithm}
\usepackage{algpseudocode}
\usepackage{amsmath,amssymb,amsthm}
\usepackage{multirow}
\usepackage{subfig}

\usepackage{makecell}

\usepackage{tikz}
\usetikzlibrary{arrows,calc,positioning,shadows,shapes}
\usepackage{pgfplots}
\pgfplotsset{compat=1.6}
\usepackage{balance}

\usepackage{lineno,hyperref}
\modulolinenumbers[5]

\hypersetup{pdfauthor={Name}}

\journal{International Journal of Information Management}

\begin{document}

\begin{frontmatter}

\title{An Investigation on Learning, Polluting, and Unlearning the Spam Emails for Lifelong Learning}



\author{Nishchal Parne}
\ead{parne\_ug@cse.nits.ac.in}
\author{Kyathi Puppaala}
\ead{kyathi\_ug@cse.nits.ac.in}
\author{Nithish Bhupathi}
\ead{bhupathi\_ug@ece.nits.ac.in}
\author{Ripon Patgiri\fnref{myfootnote}}
\ead[URL]{http://cs.nits.ac.in/rp/}
\ead{ripon@cse.nits.ac.in}
\address{National Institute of Technology Silchar}
\fntext[myfootnote]{Department of Computer Science \& Engineering, National Institute of Technology Silchar, Cachar-788010, Assam, India}




\begin{abstract}
Every activity these days generates data, from purchasing a movie ticket to researching the benefits of a product. This data is immediately trained by various models. Due to privacy, security, and usability concerns, a user may desire for their privately owned data to be forgotten by the system. Machine unlearning for security is studied in this context. Several spam email detection methods exist, each of which employs a different algorithm to detect undesired spam emails. But these models are vulnerable to attacks. Many attackers exploit the model by polluting the data, which are trained to the model in various ways. So to act deftly in such situations model needs to readily unlearn the polluted data without the need for retraining. Retraining is impractical in most cases as there is already a massive amount of data trained to the model in the past, which needs to be trained again just for removing a small amount of polluted data, which is often significantly less than 1\%. This problem can be solved by developing unlearning frameworks for all spam detection models. In this research, unlearning module is integrated into spam detection models that are based on  Naive Bayes, Decision trees, and Random Forests algorithms. To assess the benefits of unlearning over retraining, three spam detection models are polluted and exploited by taking attackers' positions and proving models' vulnerability. Reduction in accuracy and true positive rates are shown in each case showing the effect of pollution on models. Then unlearning modules are integrated into the models, and polluted data is unlearned; on testing the models after unlearning, restoration of performance is seen. Also, unlearning and retraining times are compared with different pollution data sizes on all models. On analyzing the findings, it can be concluded that unlearning is considerably superior to retraining. Results show that unlearning is fast, easy to implement, easy to use, and effective.
\end{abstract}

\begin{keyword}
Machine Learning\sep Machine Unlearning\sep Random Forest\sep Decision Trees\sep Spam Email\sep Spam Detection\sep Security\sep Retraining\sep Relearning.
\end{keyword}

\end{frontmatter}


\section{Introduction}
Machine Unlearning \cite{Cao,Neel,Brophy,Bourtoule,Aldaghri,Schelter,Du,Nguyen} is an emerging technique to unlearn the polluted data without retraining. Retraining is costlier than the unlearning. The problem nowadays is that more and more data is being generated, which results in even more data being generated, thus causing a complex data propagation network known as the lineage of the data as stated by Cao and Yang \cite{Cao}. Personal photographs, business documents, emails and website visit logs are only a few examples of the many types of data produced today by the numerous technologies in use. This information is subjected to a range of computations, including mathematical operations, to produce more data. A backup system's purpose is to protect an individual's data by duplicating it from one place to another. When images are uploaded to image storage systems, they are re-encoded in several formats and sizes to suit the system's requirements. By using the potential of advanced analytics, organizations can turn raw data, such as click logs, into useful information. This is accomplished via the use of machine learning techniques that use complex algorithms to build accurate models and features from training data. Utilizing the produced data, the next step is to develop more data, such as a movie rating prediction system that assigns ratings based on similar films. Numerous procedures use various computations, in which raw data is passed through calculation processes, expressing itself in several places and assuming a variety of forms \cite{BigData}. A sophisticated data transmission network, referred to as the lineage of the data, is constructed using a combination of data, calculations, and derived data \cite{RP}. Therefore, anomaly detection plays vital roles in modern security systems \cite{rp2,Anomaly,RP1,RP3}. In addition, detecting and protecting the spam email poses a great challenge \cite{Makkar}. Moreover, there are diverse spammer that pollutes the systems \cite{Chen,Spam}.

\subsection{Motivation}
Systems should be designed with the basic idea of completely and quickly forgetting sensitive data and its origin to restore privacy, security, and usability \cite{Cao}. Clearing systems should not remember data, even when used in a statistical process or machine learning. Clearing systems should maintain the data provenance, making this information accessible to users. It is possible to determine precisely what kind of data users wish to forget by providing the users with the option to choose degrees of granularity. Let's say a privacy-conscious user types anything such as "shameful behavior" or "inappropriate sexual behavior" in the search box while she is not(she wants to expose her identity) trying to hide her identity. In this case, the search engine could request that the search record be deleted from its database. Once the systems are done erasing the data, they go back and remove the consequences of the data, making everything from that point on seem as if the data had never been there. The members of the lineage, whether they be found online or offline, collaborate to delete data in order to transcend the confines of a system boundary. If the internet community collectively forgets, this can spread throughout the entire internet. Consumers think that suppliers with strong incentives to comply are likely to comply with customers' requests to forget and that these companies can alternatively trust models. The effectiveness of a forgetting system can be evaluated by utilizing two metrics: the overall completeness of knowledge that has been forgotten and the speed at which it can be forgotten (timeliness). The higher the solutions are scored, the more successful they are in restoring privacy, security, and usability.

A large number of prior research have shown that forgetting systems is beneficial \cite{Cao,Neel,Brophy,Bourtoule,Aldaghri,Schelter,Du,Nguyen}. A user's fundamental information can be forgotten by systems such as Google Search, but the lineage of that information is ignored. Secure deletion protects deleted data from being recovered from storage media, although it does not consider the data provenance. Information flow control can be used by forgetting systems to keep track of data lineage. Because it is designed to avoid taint explosion, it typically only checks direct data duplication rather than statistical processing or machine learning. Differential privacy is a technique for protecting the privacy of each item in a data collection similarly and consistently by restricting access to the fuzzed statistics of the whole data set. However, this restriction is in stark contrast to today's systems, such as Facebook and Google Search, which routinely access personal data to provide accurate results after obtaining permission from billions of users.

Contrary to popular belief, achieving a satisfactory balance between utility and privacy is challenging even with state-of-the-art technologies \cite{Azad}. While forgetting systems try to restore privacy to a restricted collection of data, they are not as effective as they can be. While private information can continue to circulate, the provenance of this content inside the forgetting systems is carefully tracked and wiped entirely and quickly upon request, regardless of the source. Furthermore, this fine-grained data removal considers the privacy concerns of the user and the sensitivity of the data item being removed. Forgetting systems are designed to conform to today's trust and usage paradigms, allowing for a more realistic balance between privacy and usefulness. Cao and Yang \textit{et al.} \cite{Cao} make systems more resistant to data pollution during training. In spite of these protections (and those mentioned so far, such as differential privacy), users can still request systems to erase the data as a result of policy changes and new attacks on the processes. These criteria can only be met by systems that are capable of forgetting. 

\subsection{Contribution}
On a high level, our work contrasts with all the other works mentioned above and mentioned in Table \ref{tab1e1} as this work solved a niche problem specific to spam email detection and hard coded unlearning architectures on Naive Bayes, Decision Trees, Random Forest-based spam email detectors. Results are compared using datasets from three different sources, and they also differ in ratios of number of spam mail and number of ham mail (to generalize the situation), difference in times taken for unlearning and retraining is also calculated. On the whole, spam mail detectors are coded. Multiple ways are showed to pollute the model and demonstrate the decrease in accuracy of the models, then implemented unlearning for all three models. Restoration of the lost accuracy is also showed.

\subsection{Organization}
The article is organized as follows- Section \ref{back} explores the related works on Machine Unlearning. Section \ref{meth} demonstrates the methodology of our works on Spam mails. Datasets and methods used in the experiment are described. Section \ref{res} exhibits the experimental results on Naive Bayes, Decision Tree and Random Forests for polluted spam email unlearning. Section \ref{Dis} discusses various challenges and issues on machine unlearning. Finally, Section \ref{Con} concludes the article with suitable conclusion.

\section{Background and related works}
\label{back}
Cao and Yang \cite{Cao} brought the first work on machine unlearning established the principles of statistical query learning and offered non-adaptive and adaptive statistical query learning methods. Cao and Yang \cite{Cao} used one of the two methods to perform unlearning in each of four real-world systems. They performed unlearning on Lenskit, Zozzle, Pjscan, and OSN spam filter. This study influenced many people by demonstrating the necessity for machine unlearning and discussing its benefits. 

Neel \textit{et al.} \cite{Neel} made a new difference between strong unlearning algorithms and weak unlearning algorithms. We need that the run-time of the update operation is constant (or at most logarithmic) in the length of the update sequence for an unlearning method to be strong for a given accuracy goal. The run-time per update (or, equivalently, error) of a weak unlearning algorithm can increase polynomially with the length of the update sequence. All previous research has resulted in ineffective unlearning algorithms. To remove previously learned data from models, they utilized the Gradient Descent method.

Brophy and Lowd \cite{Brophy} introduced data removal-enabled (DaRE) forests, a variant of random forests that enables the removal of training data with minimal retraining. To make data deletion efficient, DaRE trees utilize randomization and caching. Random nodes are used in the higher layers of DaRE trees, which select split attributes and thresholds evenly at random. Because they rely on the data relatively infrequently, these nodes seldom need changes. Splits are selected at lower levels to greedily optimize a split criterion such as the Gini index or mutual information. Only the required subtrees are updated when data is deleted in DaRE trees, which cache statistics and training data at each node and leaf. In the case of numerical characteristics, greedy nodes optimize across a random selection of thresholds to keep statistics while approaching the optimum threshold. DaRE trees can have a trade-off between more accurate predictions and more efficient updates by changing the number of thresholds evaluated for greedy nodes and the number of random nodes. They found that DaRE trees erase data orders of magnitude quicker than retraining from scratch while losing little to no predictive value in tests on 13 real-world datasets and one synthetic dataset.

Bourtoule \textit{et al.} \cite{Bourtoule} presented the most current study on Machine Unlearning using SISA training and the researchers are able to accelerate the unlearning process by deliberately restricting the impact of a single data point during the training method. In stateful algorithms such as stochastic gradient descent for deep neural networks, it is intended to produce the most significant possible increase in performance. When used in conjunction with transfer learning, SISA training also results in a 1.36-fold increase in the pace of retraining for complicated learning tasks such as ImageNet classification; however, accuracy is somewhat reduced as a consequence of this.

Aldaghri \textit{et al.} \cite{Aldaghri} accomplishes unlearning in all Machine Learning algorithms using ensemble learning techniques. Ensemble learning allows for the training data to be divided into smaller discontinuous shards that are then allocated to non-communicating weak learners instead of individual learning. Each shard is used to produce a weak model. The results of these models are then combined to form the final core model. This configuration presents an intrinsic trade-off between performance and unlearning cost since decreasing the shard size lowers the unlearning cost but it declines in overall performance as a result. Aldaghri \textit{et al.} \cite{Aldaghri} article proposes a coded learning protocol in which linear encoders are used to encode training data into chunks before the learning phase, then applied to the learning data. It was also shown that the related unlearning procedure meets the perfect unlearning criteria, which the researchers gave.

Schelter \textit{et al.} \cite{Schelter} present HedgeCut which is a classification model built on an ensemble of randomized decision trees that are intended to respond to unlearning requests with minimal latency.  They went into depth on how to effectively implement HedgeCut with vectorized operators for decision tree learning in a decision tree environment. Schelter \textit{et al.} \cite{Schelter} experimental evaluation on five privacy-sensitive datasets revealed that HedgeCut could unlearn training samples with a latency of around 100 microseconds and respond to up to 36,000 prediction requests per second while providing a training time and predictive accuracy that are comparable to widely used implementations of tree-based ML models such as Random Forests, according to the researchers.

Du \textit{et al.} \cite{Du} investigate the issue of lifetime anomaly detection and to suggest new methods to deal with the associated difficulties that arise. Du \textit{et al.} \cite{Du} research focused on developing a framework known as unlearning, which can be used to successfully correct the model when it is classified as false negative (or false positive). To do this, they devised several new methods to deal with two problems known as explosive loss and catastrophic forgetting, respectively. Furthermore, they developed a theoretical framework based on generative models and applied it to Du's \textit{et al.} \cite{Du} research. A general method to unlearning can be provided in this framework, and it can be used to the vast majority of zero-positive deep learning-based anomaly detection algorithms to transform those algorithms into equivalent lifetime anomaly detection solutions. They tested their method on two state-of-the-art zero-positive deep learning anomaly detection architectures as well as three real-world tasks to see if it was effective. The findings demonstrate that the suggested approach can reduce the number of false positives and false negatives by a considerable margin via the process of unlearning.

\begin{table*}[!ht]
    \centering
    \caption{Description of previous works done on Machine Unlearning}
    \begin{tabular}{|p{1.5cm}|p{7.5cm}|p{7.5cm}|}
    \hline
      \textbf{Paper}   & \textbf{Features} & \textbf{Remarks}   \\ \hline
      Cao and Yang \cite{Cao}  & \makecell[l]{1. Introduced the concept of statistical query learning.\\2. Demonstrated when to use Adaptive and\\ Non-Adaptive Statistical Query learning.} & \makecell[l]{1. Couldn't unlearn the data in the whole lineage.\\2. First study which demonstrated the necessity \\of unlearning.} \\ \hline 
      
      Neel \textit{et al.} \cite{Neel} & \makecell[l]{1. Introduced the concept of strong and weak \\machine unlearning \\ 2. Used Gradient Descent method to unlearn\\previously trained data.} & \makecell[l]{1. This is the most accurate unlearning method on\\ its release.} \\ \hline
      Brophy and Lowd \cite{Brophy} & \makecell[l]{1. It presents Data Removal-Enabled (DaRE) Random\\ Forests a variant of random forests to implement\\ machine unlearning.\\2. Tested on 13 real world datasets and 1 synthetic\\ dataset.} & \makecell[l]{1. Adept use of randomization and caching to make \\unlearning efficient.} \\ \hline
      Bourtoule \textit{et al.} \cite{Bourtoule} & \makecell[l]{1. Most recent work on unlearning which employes \\SISA training  which can forget data in \\deep neural networks.} & \makecell[l]{1. When used in conjunction with transfer learning\\ performance increases by a lot.} \\ \hline
      Aldaghri \textit{et al.} \cite{Aldaghri} & \makecell[l]{1. Training data is split into small groups and each \\group are trained to a model and ensemble of all\\ these weak models generates a strong model.} & \makecell[l]{1. Performance and unlearing cost depends on the \\size of individual shards which are part of ensemble.} \\ \hline
      Schelter \textit{et al.} \cite{Schelter} & \makecell[l]{1. Introduced Hedgecut classification model which \\responds to unlearning requests with low latency. \\2. It consist of ensemble of randomised decision trees.} & \makecell[l]{1. HedgeCut can unlearn training samples with a\\ latency of around 100 microseconds keeping \\the accuracy managable.} \\ \hline
      Du \textit{et al.} \cite{Du} & \makecell[l]{1. This study is the first one to deal with the issue \\of lifetime anomaly detection and suggested solutions. \\2. Developed solutions for explosive loss and\\ catastrophic forgetting.} & \makecell[l]{1. Model was tested on two state-of-the-art \\zero-positive deep learning anomaly detection\\ architectures and three real-world tasks.} \\ \hline
    \end{tabular}
    \label{tab1e1}
\end{table*}

\section{Methodology} 
\label{meth}
As part of this study, three algorithms are developed: Naive Bayes, Decision Trees \cite{Tan}, and Random Forests, the models are trained, polluted, and unlearned on all three datasets stated in the Dataset section. In this section, we explore how a model is trained, how an attacker can exploit the model, and how to unlearn the contaminated data on each of the above-mentioned algorithms.

\subsection{Spam Email Dataset} 
Three distinct kinds of datasets are used from three separate sources to generalise the findings. Dataset 1 \cite{Data1}, Dataset 2 \cite{Data2}, and Dataset 3 \cite{Data3} have unprocessed dataset sizes of 488 KB, 8777 KB, and 51220 KB, respectively. All the Dataset are available in Kaggle. 

\begin{table}[!ht]
\caption{Information of spam Email datasets which were used for training and testing the models before and after unlearning }
\label{table2}
\begin{tabular}{|p{1.5cm}|p{1.5cm}|p{1.5cm}|p{1cm}|p{1cm}|}
\hline
\textbf{Dataset} & \textbf{Spam mail} & \textbf{Ham mail} & \textbf{Size} & \textbf{Counts} \\ \hline
Dataset 1 \cite{Data1}  & 13.4\% & 86.6\%  & 488 KB & 5572 \\ \hline
Dataset 2 \cite{Data2} & 23.9\% & 76.1\% &  8777 KB & 5728 \\ \hline
Dataset 3 \cite{Data3} & 50.9\% & 49.1\% & 51220 KB & 33701 \\ \hline
\end{tabular}
\end{table}

In Case of Dataset 1 \cite{Data1}, a collection of 425 SMS spam messages was manually extracted from the Grumbletext Web site. This is a UK forum where mobile phone users make public accusations about SMS spam messages, with the majority of the users fail to report the spam message they received. A subset of 3,375 SMS ham messages from the NUS SMS Corpus (NSC), a collection of approximately 10,000 genuine messages gathered for research at the National University of Singapore's Department of Computer Science. The messages are mainly from Singaporeans, with the majority coming from University students. Caroline Tag's PhD Thesis yielded a collection of 450 SMS ham messages. The SMS Spam Corpus v.0.1 Big is also included containing 1,002 ham SMS messages and 322 spam messages in it. As indicated in the dataset table, this dataset contains 13.4\% spam messages and 86.6\% ham mails, resulting in findings with a higher proportion of ham mails and less spam mails. 

In the case of Dataset 2 \cite{Data2}, the original source is unclear, but it is used as training data in a Kaggle competition for identifying spam emails. It includes 23.8 percent spam emails and 76.2 percent ham emails. In the instance of Dataset 3 \cite{Data3}, the same dataset is used, which is used in \cite{metsis2006spam}. It has about equal amounts of spam and ham emails. This is the biggest of the three datasets, comprising 33701 emails. A custom dataset of spam emails is generated from different online shopping sites for Decision trees just to demonstrate the kind of pollution employed in Decision trees.

\subsection{Naive Bayes}
Naive Bayes is a machine learning algorithm that employs the Bayes Theorem. Because of its simplicity and efficacy, Naive Bayes is often employed in text categorization applications and experiments. The chi-squared test is used to select features as it helps in selecting features which are responsible for classifying a mail as spam and exclude other independent features which doesn't affect the probability of a mail being spam. It predicts membership probabilities for every class, i.e the likelihood that a given record or data point belongs to a particular class. The class with the highest probability is regarded the more likely class; in our example, spam and ham are classes of mails which are to be predicted.

This form of Naive Bayes is known as Multinomial Naive Bayes.
To enable unlearning in Naive Bayes, the model is designed in such a way that it can be trained incrementally subsequently, it can be unlearned incrementally. To do this, a new class is created which includes the unlearn function along with fundamental methods such as fit and predict.

In the Naive Bayes model, incremental characteristics are trivial. To make the original algorithm incremental, just a few modifications are required. We should simply keep the data in memory and update the counts based on the data encountered. After each training phase, the Chi-squared test is performed, resulting in a change in selected features after each training phase.

\subsubsection{Model Pollution}
An attacker can create spam mails by introducing traits and these are absent in any ham mails in order to perform data pollution and influence the detection of Naive Bayes model. In this instance, the constructed spam samples mails can be predicted as spam by the model and included in the training data set. The injected features in the crafted spam mails can impact the model's feature selection as well as the sample detection stage.

\begin{equation}
\begin{split}
    chi^{2}=\frac{f_1()}{f_2()}
\end{split}
\end{equation}
where
\begin{equation*}
\begin{split}
    f_1()=&(Nsf\ Nh\hat{f}\ +\ Ns\hat{f}\ Nhf)^{2}\\
    f_2()=&(Nsf+ Ns\hat{f})(Nhf+ Nh\hat{f})\\&(Nsf+Nhf)(Ns\hat{f}\ +Nh\hat{f})
\end{split}
\end{equation*}

In feature selection stage model selects highly dependent features for training the model using chi-squared test. chi-squared test equation comprises of two functions f\textsubscript{1}() and f\textsubscript{2}() where $f_1()=(Nsf\ Nh\hat{f}\ +\ Ns\hat{f}\ Nhf)^{2}$ and $ f_2()=(Nsf+ Ns\hat{f})(Nhf+ Nh\hat{f})(Nsf+Nhf)(Ns\hat{f}\ +Nh\hat{f})$
Here $Nsf$ is the number of spam samples with feature f, $Nh\hat{f}$ is the number of ham samples without feature f, $Ns\hat{f}$ is the number of spam samples without feature f, $Nhf$ is the number of ham samples with feature f.

As pollution samples contains many crafted spam features it affects both feature selection as well as sample detection. In case of feature selection, a particular feature is selected if value of $chi^{2}$ exceeds 99\% confidence of feature being dependent. First, because the crafted features do not appear in ham samples but only spam samples, in the chi-squared test, $Nsf$ and $Nh\hat{f}$ are large, and $Nhf$ and $Ns\hat{f}$ are small. Therefore, the feature selection process is likely to pick the crafted pollution features compared to actual spam features. In addition, the injected features can make a real spam feature that would have been selected less likely to be selected. Because the attacker does not change any ham training sample or remove real features existing malicious samples, but only add new samples, in the chi-squared test, So for a real spam feature $Nsf$ , $Nhf$ and $Nh\hat{f}$ remain the same, and $Ns\hat{f}$ increases. Therefore, the feature selection process is less likely to pick up real spam feature which decreases the true positive rate significantly. Second, the presence of an injected feature lowers the accuracy of the naive Bayes classifier. Because the classifier splits the weight of a mail being spam between the presence of real and crafted features. 

\subsubsection{Unlearning}
The following is how unlearning works in Naive Bayes. First, the unlearning process aggregates all of the data to be forgotten and extracts relevant characteristics from the data to be forgotten. The chi values of all the characteristics are then updated. Because $Nsf$, $Nhf$, $Ns\hat{f}$, and $Nh\hat{f}$ in the chi value computation are counts of samples or a sum of outputs of indicator functions, the chi value of features can be readily updated. If a feature's chi value falls below a certain level, it is eliminated. The updated chi values are then used to produce a fresh list of features. Second, the unlearning procedure updates all of the conditional probability values associated with the newly discovered features in the first phase. Because none of the aforementioned changes entail increasing the size of the training dataset, the time complexity is O(q), where q is the number of features. Naive Bayes classifiers can forget 100\% of the polluted data samples,unlearned model performs exactly similar to the model without pollution data.

\subsection{Decision Trees}
A  decision  tree  is  a  type of machine learning algorithm that is useful for classification or regression. Here, It is used for text classification. The purpose of a Decision Tree is to develop a training model capable of predicting the class or value of a target variable by inferring basic decision rules from previous data (training data). In this instance, the target variable is the mail's label (spam or ham). A decision tree classifies inputs by segmenting the input space into regions. Multiple methods are used by decision trees to determine whether to divide a node into two or more sub-nodes. Sub-node generation improves the homogeneity of the resulting sub-nodes. In other words, the purity of the node improves as the target variable rises. The decision tree divides the nodes into sub-nodes based on all available factors and then chooses the split that produces the most homogeneous sub-nodes.

Our aim is to train and unlearn an incremental decision tree in batches. To achieve this VFDT is used ( Very Fast Decision Tree). It is also known as Hoeffedding tree. With the influx of data, new branches are constantly being created, while old branches are being phased out. This tree is used because it is incremental in nature.

\subsubsection{Model Pollution}
To contaminate the Decision tree model, a method is employed that is diametrically opposite to the approach used to pollute Naive Bayes, in which the model is polluted using crafted spam messages. However, in this method, pollution is done using ham messages. The following points show how the model is impacted and how an attacker can benefit from it.

Considering a promotional spam mail from an online store named "ABC", the store had started email marketing campaign and sends emails to all it's users let the email as follows-
\begin{verbatim}
"Get up to 40% off on t-shirts for men on 
ABC Fashion. Press this link 
https://www.ABCfashion.com/lp/mens-t-shirt
t-shirts to grab the offer now.
Use code "HAPPY40" during checkout"
\end{verbatim}

As a result of the use of terms like "off," "link," "checkout," and "code," the email marketing model classifies this message as spam. As a result, the attacker tries to deceive the model by making the store's name sound hammy by using the word in ham mails. The Decision Tree, as previously stated, seeks to partition the sub trees in the most homogeneous way possible after training the model using ham mails that contain the shop name. As a result, during training, a subtree is constructed with the store name as a decision, indicating that if the store word appears in a mail, the letter is most likely a ham mail. It is beneficial so that this email does not land up in the spam bin for future marketing attempts. This email, on the other hand, is spam, and it has no bearing on the user. This allows the attacker to be identified who is manipulating the model and modifying the results to enhance the store's email marketing efforts. We'll examine,how to unlearn the polluted mails in Decision Trees in this part.

\subsubsection{Unlearning}
VFDT, which is based on the improvement of hoeffding trees, is used to dynamically generate the decision tree. Because the model can train in batches and construct and update subtrees dynamically during training, we train the polluted data with labels that are opposite to the model (which takes the least amount of time). As a result of training the polluted data with the opposite label, the subtree's homogeneity is broken, the choice is completely eliminated, and the polluted data has no effect on subsequent label predictions. So, training the model with opposite labels to VFDT we can inculcate unlearning in Decision Trees.

\subsection{Random Forests}
The random forest is a classification method made up of multiple decision trees. It employs bagging and feature randomization in the construction of each individual tree in an attempt to generate an uncorrelated forest of trees whose forecast by committee is more accurate than that of any individual tree. Because decision trees are highly sensitive to the data on which they are trained, they are prone to overfitting. Random forest, on the other hand, takes advantage of this problem by allowing each tree to randomly select from the dataset in order to get diverse tree topologies. Some distinctions exist between random forest and a collection of decision trees. Training dataset provides input to a decision tree and, as a result, it produce rules to generate predictions.

For example, in order to forecast whether a person would watch a video, you can gather data from prior videos watched by the user, as well as some of the videos' characteristics. A decision tree representation of characteristics and labels results in certain rules that assist in determining if a video is watched by the user. Compared to the random forest technique, which uses randomly selected observations and features to create several decision trees and then averages the results.

To enable unlearning in Random Forests,  incremental random forests are created by adding a new set of trees to the forest set at each training step. The number of trees added at each training phase is determined by the size of the training data. So, as part of the study, we attempted to contaminate a random forest-based spam mail detection model and demonstrate the reduction in model performance, as well as offer a technique to unlearn the polluted data without retraining the model from start. 

\subsubsection{Model Pollution}
In the case of Random forests, the third method of contaminating a model is used: human interference. A person gives many of the models their initial training set manually. In this instance, the training labels can be manually changed by a person for a variety of reasons. For instance, if the model is trained with actual spam mail but with a ham label of 0 in our case, and vice versa, this human participation has a significant impact on the model's performance.
Because the model hasn't been trained yet and is given contaminated data, it makes incorrect conclusions. For example, if a spam email is sent with the word "Free" in it and label it with 0 (ham label), the model would be prone to predicting the next mail with the phrase "Free" as ham, which is inaccurate and leads to a drop in accuracy.  

\subsubsection{Unlearning}
As stated earlier, we polluted random forests model using human interference method, so to unlearn the polluted model the model needs to be incremental. To negate the impact of previously trained trees, additional trees are added to the forest set and those trees are trained with complement labels after making the model incremental. 

Consider the preceding example, in which the word "Free," which is a genuine spam term, was taught to the model as ham. To unlearn it,  additional trees are added to the forests set and trained those extra trees the word "Free" as ham. Half of the trees predict the word as spam, while the other half predicts it as ham, resulting in the term "Free" being forgotten as a ham word. Number of trees to added depends on various factors like previous forest size, scale at which word should be unlearned which is decided by the model while unlearning. 

Random forests is more robust to pollution than other machine learning algorithms because with a few polluted mails, effect is unnoticed as only few trees predict the wrong label while the majority predicts the correct label. So to pollute the model greater percent of pollution data is required to alter the results.

\section{Experimental Results and Analysis}
\label{res}
As previously said training, polluting, and unlearning are done on all the three datasets listed in the datasets section. Results and analysis on how model performed before, after pollution and after unlearning is discussed for each algorithm below.  

\subsection{Naive Bayes}
\begin{figure}[!ht]
\centering
\begin{tikzpicture}
    \begin{axis}[
          ybar,
            bar width=.1cm,
            width=0.5\textwidth,
            height=.2\textwidth,
            enlarge x limits=0.15,
            legend style={at={(0.5,1)},
                anchor=south,legend columns=4,legend cell align=left},
            symbolic x coords={Before pollution,After pollution,After unlearning},
            xtick=data,
             x tick label style={anchor=north},
            nodes near coords align={vertical},
            ymin=0,ymax=100,
            ylabel={accuracy},xlabel={Training stage}
        ]
        \addplot table[x=interval,y=Dataset 1]{\nb};
        \addplot table[x=interval,y=Dataset 2]{\nb};
        \addplot table[x=interval,y=Dataset 3]{\nb};
        \legend{Dataset 1, Dataset 2, Dataset 3}
    \end{axis}
\end{tikzpicture}
\caption{Accuracy for Naive Bayes algorithm before pollution, after pollution and after unlearning}
\label{figure1}
\end{figure}
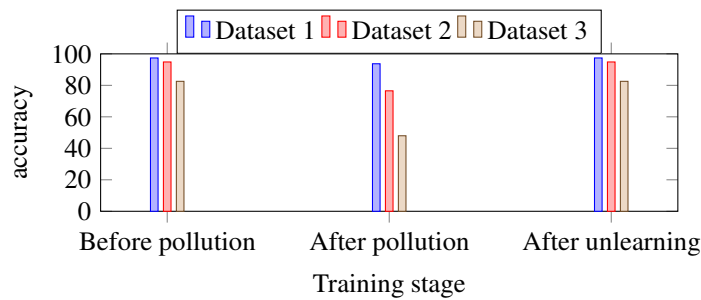

 Good accuracy is attained after training the model with 80\% of the dataset and testing it with other 20\%. Accuracies are 97.4\%, 94.8\%, 82.5\% for Dataset 1, Dataset 2, and Dataset 3 respectively. Then a pollution dataset is crafted and labels of crafted samples are predicted and the models are trained with the predicted values. Which affected the models accuracy significantly with just 5 percent polluted data. Then, as described in the methodology section, unlearning strategy is applied. Because the Naive Bayes algorithm can completely unlearn contaminated data, the accuracies before and after pollution are the same. Fig. \ref{figure1} depicts the accuracy of Naive Bayes at various training phases.

Less change in accuracy can be seen in Dataset 1 \cite{Data1} and Dataset 2 \cite{Data2} compared to Dataset 3 \cite{Data3} because pollution only affects the true positive rates of the models, which is the attacker's main intention (actual spam mails must be predicted as ham), True positive rates can be seen in Fig. \ref{figure2}. True positive rates depicts the probability of classifying a spam mail as spam, here in case of dataset 1 there are less than 20\% spam mails so after pollution only these 20\% datapoints are affected, resulting in less change in accuracy; however, a significant decrease in true positive rates of the models is observed, as shown in Figure 2. True positive rates have decreased to almost nil for Dataset 2 and Dataset 3, and to 35\% for Dataset 1. Similarly True negative rates are shown in Fig. \ref{figure3}

Regarding the completeness is concerned, in case of Naivebayes polluted data is 100\% unlearned, i.e., model performs as if it had never seen the polluted data samples after unlearning , resulting in the same accuracy, true positive rates and true negative rates before pollution and after unlearning.

On examining the times required for unlearning and retraining the model, it is found that unlearning provides the most advantage, as shown in Table \ref{table3} . A dataset with 10000 mails is used and both unlearning and retraining are performed on it. A significant difference is  observed in Naive Bayes between these times taken. The time difference decreases as the size of the sample to be unlearned increases; however, in most real-world instances, the sample to be unlearned is far less than 1\% of the training data. The time difference also grows in proportion to the amount of the training data. 

It is noticed that unlearning a single contaminated mail from 10000 mails takes about 0.001 sec, while retraining the model without including that mail takes 160sec, which is inadequate for unlearning a single mail. Also in cases with 1\%, 10\%, 30\% pollution data times taken to retrain the model are 158 sec, 147 sec, 117 sec whereas times taken to unlearn the same data are 2.25 sec, 18.75 sec, 53.67 sec. All these cases unlearning outperforms retraining. Results of time disparities between unlearning and retraining in Naive Bayes is shown in Table 1.

 With all of these data, unlearning clearly outperforms retraining in every way. The only occasion where the time for unlearning exceeds the time for retraining in Naive Bayes is when the sample to be retrained is large, i.e., more than 50\% of the dataset, which is uncommon in most circumstances.

\pgfplotstableread[row sep=\\,col sep=&]{
interval& Dataset 1&  Dataset 2& Dataset 3\\
Before pollution&	97.39&	94.85&	82.55\\
After pollution&	93.71&	76.52&	47.98\\
After unlearning&	97.39&	94.85&	82.55\\
}\nb

\pgfplotstableread[row sep=\\,col sep=&]{
interval& Dataset 1&  Dataset 2& Dataset 3\\
Before pollution&	96.22&	80.00&	67.20\\
After pollution&	56.60&	0.37&	0.27\\
After unlearning&	96.22&	80.00&	67.20\\
}\nbs
\begin{figure}[!ht]
\centering
\begin{tikzpicture}
    \begin{axis}[
          ybar,
            bar width=.1cm,
            width=0.5\textwidth,
            height=.2\textwidth,
            enlarge x limits=0.15,
            legend style={at={(0.5,1)},
                anchor=south,legend columns=4,legend cell align=left},
            symbolic x coords={Before pollution,After pollution,After unlearning},
            xtick=data,
             x tick label style={anchor=north},
            nodes near coords align={vertical},
            ymin=0.01,ymax=100,
            ylabel={True positive rate},xlabel={Training stage},
        ]
        \addplot table[x=interval,y=Dataset 1]{\nbs};
        \addplot table[x=interval,y=Dataset 2]{\nbs};
        \addplot table[x=interval,y=Dataset 3]{\nbs};
        \legend{Dataset 1, Dataset 2, Dataset 3}
    \end{axis}
\end{tikzpicture}
\caption{True positive rates for Naivebayes algorithm before pollution, after pollution and after unlearning}
\label{figure2}
\end{figure}

\pgfplotstableread[row sep=\\,col sep=&]{
interval& Dataset 1&  Dataset 2& Dataset 3\\
Before pollution&	97.59&	99.42&	99.19\\
After pollution&	99.89&	100.00&	99.69\\
After unlearning&	97.59&	99.42&	99.19\\
}\nbs
\begin{figure}[!ht]
\centering
\begin{tikzpicture}
    \begin{axis}[
          ybar,
            bar width=.1cm,
            width=0.5\textwidth,
            height=.2\textwidth,
            enlarge x limits=0.15,
            legend style={at={(0.5,1)},
                anchor=south,legend columns=4,legend cell align=left},
            symbolic x coords={Before pollution,After pollution,After unlearning},
            xtick=data,
             x tick label style={anchor=north},
            nodes near coords align={vertical},
            ymin=96,ymax=101,
            ylabel={True Negative rate},xlabel={Training stage},
        ]
        \addplot table[x=interval,y=Dataset 1]{\nbs};
        \addplot table[x=interval,y=Dataset 2]{\nbs};
        \addplot table[x=interval,y=Dataset 3]{\nbs};
        \legend{Dataset 1, Dataset 2, Dataset 3}
    \end{axis}
\end{tikzpicture}
\caption{True negative rates for Naivebayes algorithm before pollution, after pollution and after unlearning}
\label{figure3}
\end{figure}

\begin{table}[!ht]
\caption{Comparison between times taken while retriaining and unlearning data in Naivebayes on 10000 mails of training data}
\label{table3}
\begin{tabular}{|p{1.5cm}|c|c|c|}
\hline
\textbf{Unlearning size} & \textbf{Retrain speed} & \textbf{unlearn speed} & \multicolumn{1}{l|}{\textbf{Speed up}} \\ \hline
1 mail           & 160 sec & 0.001 sec & 160000 x \\ \hline
1\%    & 158 sec & 2.25 sec  & 70.2 x   \\ \hline
10\% & 147 sec & 18.75 sec & 7.86 x   \\ \hline
30\% & 117 sec & 53.67 sec & 2.17 x   \\ \hline
\end{tabular}
\end{table}

\subsection{Decision Trees}
As previously stated, different type of pollution is used in Decision trees, which differs from the technique used in Naive Bayes. By altering the model, the attacker attempted to make his promotional spam emails ham. When the homogeneity of a decision tree is substantially changed, the accuracy decreases abruptly as the decision making node changes.

\begin{figure}[!ht]
\centering
\begin{tikzpicture}
    \begin{axis}[
          ybar,
            bar width=.1cm,
            width=0.45\textwidth,
            height=.2\textwidth,
            enlarge x limits=0.15,
            legend style={at={(0.5,1)},
                anchor=south,legend columns=4,legend cell align=left},
            symbolic x coords={Before pollution,After pollution,After unlearning},
            xtick=data,
             x tick label style={anchor=north},
            nodes near coords align={vertical},
            ymin=45,ymax=102,
            ylabel={accuracy},xlabel={Training stage}
        ]
        \addplot table[x=interval,y=Dataset1]{\dt};
        \addplot table[x=interval,y=Dataset2]{\dt};
        \addplot table[x=interval,y=Dataset3]{\dt};
        \legend{Dataset 1, Dataset 2, Dataset 3}
    \end{axis}
\end{tikzpicture}
\caption{Accuracy for Decision Tree algorithm before pollution, after pollution and after unlearning}
\label{figure4}
\end{figure}
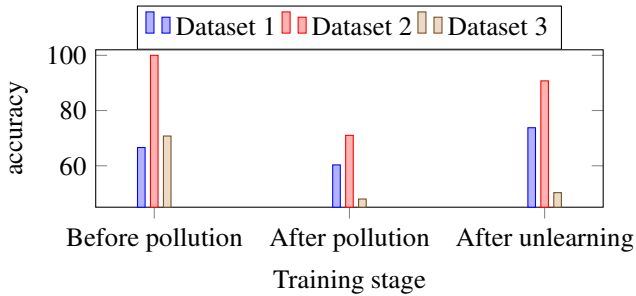

After training the model with same datasets which are used in  Naive Bayes but as we have discussed in methodology section we used different dataset for predictions which consists of promotional mails from various online stores; therefore, the sources of datasets and the prediction dataset are completely different. Top quality prediction accuracies of the models can't be expected and  as shown in Figure \ref{figure4}, before pollution are 66.61\%, 100.00\%, 70.78\% for Dataset 1, Dataset 2, Dataset 3 respectively and after the pollution it dropped to 60.32\%, 71.02\%,	47.97\% where it is insignificant the decrement of accuracy in Dataset 1 and 2 as they mostly consist of ham mails, after unlearning accuracies again went to 73.78\%, 90.75\%, 50.27\%. It is clear that in Dataset 3 there is not can of a recovery in accuracy as in case of Decision trees the Decision node changes only when the entropy changes where in case of Dataset 3, the unlearning size is small as compared to original dataset; thus, the values regarding other decision nodes are way higher, resulting in less recovery of accuracy, this can be overcome by introducing parameter of unlearning magnitude to magnify the effect to a greater extent. 

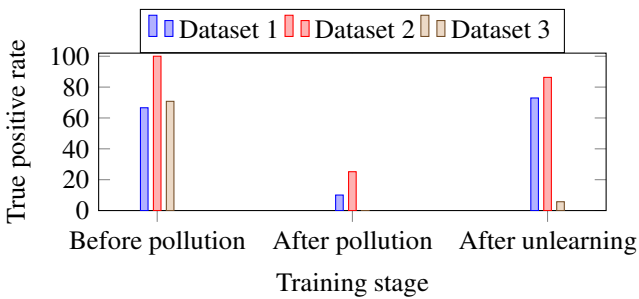
\begin{figure}[!ht]
\centering
\begin{tikzpicture}
    \begin{axis}[
          ybar,
            bar width=.1cm,
            width=0.45\textwidth,
            height=.2\textwidth,
            enlarge x limits=0.15,
            legend style={at={(0.5,1)},
                anchor=south,legend columns=4,legend cell align=left},
            symbolic x coords={Before pollution,After pollution,After unlearning},
            xtick=data,
             x tick label style={anchor=north},
            nodes near coords align={vertical},
            ymin=0,ymax=102,
            ylabel={True positive rate},xlabel={Training stage}
        ]
        \addplot table[x=interval,y=Dataset 1]{\dts};
        \addplot table[x=interval,y=Dataset 2]{\dts};
        \addplot table[x=interval,y=Dataset 3]{\dts};
        \legend{Dataset 1, Dataset 2, Dataset 3}
    \end{axis}
\end{tikzpicture}
\caption{True positive Rates for Decision Tree algorithm before pollution, after pollution and after unlearning}
\label{figure5}
\end{figure}

The true positive rates as seen in Fig. \ref{figure5}, are 66.61\%, 100.00\%, 70.78\% for Dataset 1, Dataset 2, Dataset 3 respectively, which are same as accuracies because all the mails are spam mails as each and every mail is intended to promote a product. The True positives after pollution are 10.06\%, 25.18\%, 0.02\% for Dataset 1, Dataset 2, Dataset 3 respectively which are badly affected by pollution. But after unlearning these true positive rates rise to 72.95\%, 86.29\% and 5.73\% for Dataset 1, Dataset 2, Dataset 3 respectively. 

Now Considering times taken to unlearn vs time taken to retrain as presented in Table \ref{table4}, unlearning is faster than the retraining, while unlearning it takes only 2.56 sec, 23.20 sec, and 63.85 sec to unlearn 1\%, 10\%, and 30\% polluted data as shown in table 4, whereas for retraining it takes 7600 sec, 6900 sec, 5500 sec. In most situations, the amount of data to be unlearned less than 1\%, where unlearning easily outperforms retraining.

With these results, it can be concluded that, although unlearning in Decision trees is not 100\% complete, it is most plausible and convenient to introduce unlearning in Decision trees as it takes way less time to unlearn rather than retraining the model. Also, in most of the cases, the polluted data are less than 1\% of the whole dataset which implies the unlearning benefit maximized.

\pgfplotstableread[row sep=\\,col sep=&]{
interval& Dataset1&  Dataset2& Dataset3\\
Before pollution&	66.61&	100.00&	70.78\\
After pollution&	60.32&	71.02&	47.97\\
After unlearning&	73.78&	90.75&	50.27\\
}\dt

\pgfplotstableread[row sep=\\,col sep=&]{
interval& Dataset 1&  Dataset 2& Dataset 3\\
Before pollution&	66.61&	100.00&	70.78\\
After pollution&	10.06&	25.18&	0.02\\
After unlearning&	72.95&	86.29&	5.73\\
}\dts

\begin{table}[!ht]
\caption{Comparison between times taken while retriaining and unlearning data in Decision Trees on 10000 mails of training data}
\label{table4}
\begin{tabular}{|p{1.5cm}|p{1.5cm}|p{1.5cm}|c|}
\hline
\textbf{Unlearning size} & \textbf{Retrain speed} & \textbf{unlearn speed} & \multicolumn{1}{l|}{\textbf{Speed up}} \\ \hline
1 mail           & 203 sec & 0.002 sec & 101500 x \\ \hline
1\%  & 201 sec & 2.37 sec  & 84.81 x   \\ \hline
10\%  & 185 sec & 21.61 sec &  8.56 x   \\ \hline
30\%  & 140 sec & 60.08 sec & 2.33 x   \\ \hline
\end{tabular}
\end{table}

\subsection{Random Forests}
\begin{figure}[!ht]
\centering
\begin{tikzpicture}
    \begin{axis}[
          ybar,
            bar width=.1cm,
            width=0.45\textwidth,
            height=.2\textwidth,
            enlarge x limits=0.15,
            legend style={at={(0.5,1)},
                anchor=south,legend columns=4,legend cell align=left},
            symbolic x coords={Before pollution,After pollution,After unlearning},
            xtick=data,
             x tick label style={anchor=north},
            nodes near coords align={vertical},
            ymin=87,ymax=98,
            ylabel={Accuracy},xlabel={Training stage},
        ]
        \addplot table[x=interval,y=Dataset1]{\rf};
        \addplot table[x=interval,y=Dataset2]{\rf};
        \addplot table[x=interval,y=Dataset3]{\rf};
        \legend{Dataset 1, Dataset 2, Dataset 3}
    \end{axis}
\end{tikzpicture}
\caption{Accuracy for Random Forests algorithm before pollution, after pollution and after unlearning}
\label{figure6}
\end{figure}
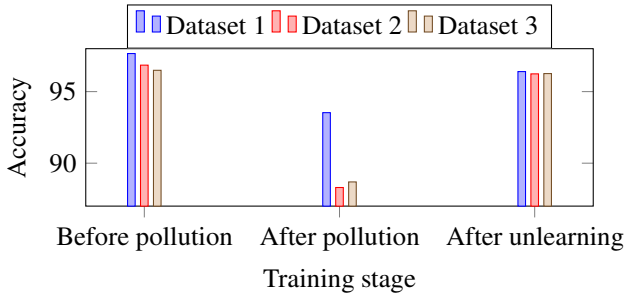

Random Forests are polluted as well, however the findings show that Random Forests are the most robust to pollution, since the entropy of more than half of the decision trees must change to alter the tree structures, resulting in the algorithm's resilience. Accuracies, as shown in Fig. \ref{figure6}, without contaminated data are 97.66\%, 96.85\%, and 96.49\% for Dataset 1, Dataset 2, and Dataset 3 respectively which is an acceptable accuracy.

When one-fifth of the data is polluted, the model's accuracies drop less than in other models after polluting the model. After polluting the model as mentioned in the methodology section, the polluted data is unlearned, which significantly improves accuracy. Accuracy after unlearning the polluted data is 96.40\%, 96.24\%, and 96.26\% for datasets 1, 2, and 3, respectively. Polluted data is incompletely unlearned, but it is slightly less than 100\%, which could be a simple trade-off for a time advantage. 

So, in order to unlearn, the number of estimators are increased and those new estimators are trained with opposite labels to unlearn a specific sample. This substantially increased accuracy. For datasets 1, 2, and 3, the accuracy after unlearning is 96.40\%, 96.24\%, and 96.26\%, respectively. Although the model unlearns the polluted data partially, it provided almost similar findings, falling just short of 100 percent. The degree of unlearning can also be changed by altering the number of additional estimators to be added to the model at each unlearning session.

Although there is indifference in accuracies before and after pollution, even minor gains can make a dramatic impact in large-scale applications. It is observed that Random forest is more resistant to pollution, thus it takes more polluted data to influence the model than that of other algorithms. 

On considering the time disparities between retraining and unlearning, the actual advantage of unlearning becomes clear. When retraining the model without polluted data, the model takes 1.15 seconds, 1.10 seconds, and 0.84 seconds with 1\%, 10\%, and 30\% polluted data respectively. Whereas in case of unlearning it takes only 0.03 sec, 0.12 sec, and 0.32 sec to unlearn the same 1\%, 10\%, and 30\% polluted data as shown in table 3. In most situations, the amount of data to be unlearned less than 1\%, where unlearning easily outperforms retraining.

\pgfplotstableread[row sep=\\,col sep=&]{
interval& Dataset1&  Dataset2& Dataset3\\
Before pollution&	97.66&	96.85&	96.49\\
After pollution&	93.53&	88.30&	88.69\\
After unlearning&	96.40&	96.24&	96.26\\
}\rf

\pgfplotstableread[row sep=\\,col sep=&]{
interval& Dataset1&  Dataset2& Dataset3\\
Before pollution&	87.42&	91.85&	96.23\\
After pollution&	76.10&	77.77&	86.08\\
After unlearning&	86.24&	91.85&	96.97\\
}\rf
\begin{figure}[!ht]
\centering
\begin{tikzpicture}
    \begin{axis}[
          ybar,
            bar width=.1cm,
            width=0.45\textwidth,
            height=.2\textwidth,
            enlarge x limits=0.15,
            legend style={at={(0.5,1)},
                anchor=south,legend columns=4,legend cell align=left},
            symbolic x coords={Before pollution,After pollution,After unlearning},
            xtick=data,
             x tick label style={anchor=north},
            nodes near coords align={vertical},
            ymin=75,ymax=98,
            ylabel={True Positive rate},xlabel={Training stage},
        ]
        \addplot table[x=interval,y=Dataset1]{\rf};
        \addplot table[x=interval,y=Dataset2]{\rf};
        \addplot table[x=interval,y=Dataset3]{\rf};
        \legend{Dataset 1, Dataset 2, Dataset 3}
    \end{axis}
\end{tikzpicture}
\caption{True Positive Rates for Random Forests algorithm before pollution, after pollution and after unlearning}
\label{figure7}
\end{figure}
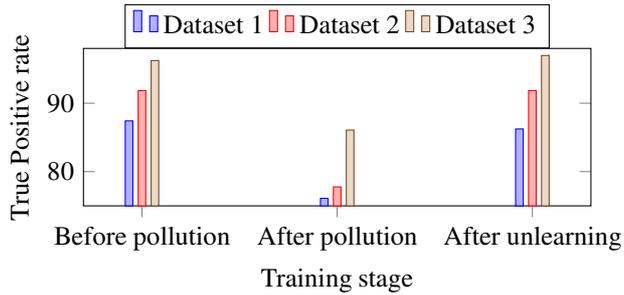

\pgfplotstableread[row sep=\\,col sep=&]{
interval& Dataset1&  Dataset2& Dataset3\\
Before pollution&	99.47&	98.40&	96.66\\
After pollution&	97.38&	92.46&	90.87\\
After unlearning&	99.16&	98.40&	95.73\\
}\rf
\begin{figure}[!ht]
\centering
\begin{tikzpicture}
    \begin{axis}[
          ybar,
            bar width=.1cm,
            width=0.45\textwidth,
            height=.2\textwidth,
            enlarge x limits=0.15,
            legend style={at={(0.5,1)},
                anchor=south,legend columns=4,legend cell align=left},
            symbolic x coords={Before pollution,After pollution,After unlearning},
            xtick=data,
             x tick label style={anchor=north},
            nodes near coords align={vertical},
            ymin=87,ymax=100,
            ylabel={True Negative Rate},xlabel={Training stage},
        ]
        \addplot table[x=interval,y=Dataset1]{\rf};
        \addplot table[x=interval,y=Dataset2]{\rf};
        \addplot table[x=interval,y=Dataset3]{\rf};
        \legend{Dataset 1, Dataset 2, Dataset 3}
    \end{axis}
\end{tikzpicture}
\caption{True Negative Rate for Random Forests algorithm before pollution, after pollution and after unlearning}
\label{figure8}
\end{figure}
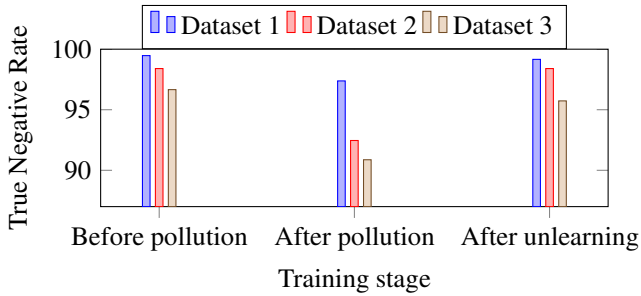

\begin{table}[!ht]
\caption{Comparison between times taken while retraining and unlearning data in Random Forests on 10000 mails of training data}
\label{table5}
\begin{tabular}{|p{1.5cm}|c|c|c|}
\hline
\textbf{Unlearning size} & \textbf{Retrain speed} & \textbf{unlearn speed} & \multicolumn{1}{l|}{\textbf{Speed up}} \\ \hline
1\%     & 1.15 sec & 0.03 sec & 38.3 x \\ \hline
10\%   & 1.10 sec & 0.12 sec  & 9.17 x   \\ \hline
30\%  & 0.84 sec & 0.32 sec &  2.62 x   \\ \hline
\end{tabular}
\end{table}

\subsection{Overall Accuracy}

\pgfplotstableread[row sep=\\,col sep=&]{
interval& Dataset1&  Dataset2& Dataset3\\
Naive Bayes&	97.39&	94.85&	82.55\\
Decision Tree&	66.61&	100&	70.78\\
Random Forest&	97.66&	96.85&	96.49\\
}\ev
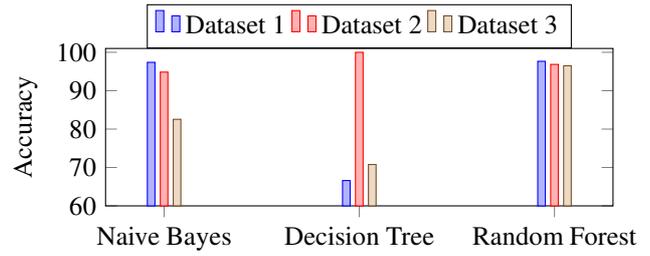
\begin{figure}[!ht]
\centering
\begin{tikzpicture}
    \begin{axis}[
          ybar,
            bar width=.1cm,
            width=0.45\textwidth,
            height=.2\textwidth,
            enlarge x limits=0.15,
            legend style={at={(0.5,1)},
                anchor=south,legend columns=3,legend cell align=left},
            symbolic x coords={Naive Bayes,Decision Tree,Random Forest},
            xtick=data,
             x tick label style={anchor=north},
            nodes near coords align={vertical},
            ymin=60,ymax=101,
            ylabel={Accuracy},
        ]
        \addplot table[x=interval,y=Dataset1]{\ev};
        \addplot table[x=interval,y=Dataset2]{\ev};
        \addplot table[x=interval,y=Dataset3]{\ev};
        \legend{Dataset 1, Dataset 2, Dataset 3}
    \end{axis}
\end{tikzpicture}
\caption{Overall evaluation of Naive Bayes, Decision Tree and Random Forest before pollution.}
\label{figure9}
\end{figure}

Figure \ref{figure9} demonstrates the accuracy before the pollution. Decision Tree achieves the highest accuracy of 100\% in Dataset2 while it achieves the lowest accuracy of 70.78\% among the three algorithms. It is observed that Random Forest achieves consistent accuracy more than 96\% in all datasets while the accuracy of Naive Bayes algorithms fluctuate in Dataset2 and Dataset3. This accuracy is observed before polluting the algorithms.

\pgfplotstableread[row sep=\\,col sep=&]{
interval& Dataset1&  Dataset2& Dataset3\\
Naive Bayes&	93.71&	76.52&	47.98\\
Decision Tree&	60.32&	71.02&	47.97\\
Random Forest&	93.53&	88.3&	88.69\\
}\eva
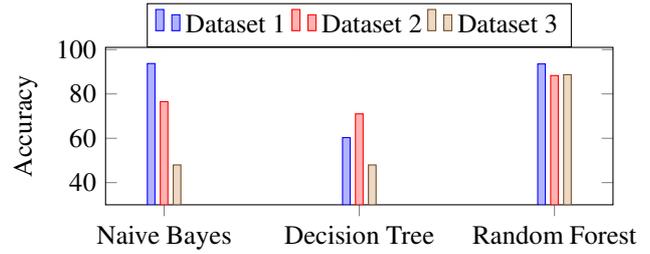
\begin{figure}[!ht]
\centering
\begin{tikzpicture}
    \begin{axis}[
          ybar,
            bar width=.1cm,
            width=0.45\textwidth,
            height=.2\textwidth,
            enlarge x limits=0.15,
            legend style={at={(0.5,1)},
                anchor=south,legend columns=3,legend cell align=left},
            symbolic x coords={Naive Bayes,Decision Tree,Random Forest},
            xtick=data,
             x tick label style={anchor=north},
            nodes near coords align={vertical},
            ymin=30,ymax=101,
            ylabel={Accuracy},
        ]
        \addplot table[x=interval,y=Dataset1]{\eva};
        \addplot table[x=interval,y=Dataset2]{\eva};
        \addplot table[x=interval,y=Dataset3]{\eva};
        \legend{Dataset 1, Dataset 2, Dataset 3}
    \end{axis}
\end{tikzpicture}
\caption{Overall evaluation of Naive Bayes, Decision Tree and Random Forest after pollution.}
\label{figure10}
\end{figure}

Figure \ref{figure10} reveals the accuracy after pollution. An algorithm can easily be polluted by the adversaries. Therefore, the accuracy of the three algorithms drastically fall down which endanger entire learning systems. Figure \ref{figure10} evidence the downgrading the accuracy in pollution. The accuracy of the Naive Bayes falls to 93.71\%, 76.52\%, and 47.98\% in Dataset1, Dataset2, and Dataset3 respectively. Similarly, the accuracy of the Decision Tree falls to 60.32\%, 71.02\%	and 47.97\% in Dataset1, Dataset2, and Dataset3 respectively. Likewise, the accuracy of the Random Forest falls to 93.53\%,	88.3\%, and	88.69\% in Dataset1, Dataset2, and Dataset3 respectively. The significant falls in accuracy alarms the malfunctioning of the machine learning algorithms.

\pgfplotstableread[row sep=\\,col sep=&]{
interval& Dataset1&  Dataset2& Dataset3\\
Naive Bayes&	97.39&	94.85&	82.55\\
Decision Tree&	73.78&	90.75&	50.27\\
Random Forest&	96.4&	96.24&	96.26\\
}\evr
\begin{figure}[!ht]
\centering
\begin{tikzpicture}
    \begin{axis}[
          ybar,
            bar width=.1cm,
            width=0.45\textwidth,
            height=.2\textwidth,
            enlarge x limits=0.15,
            legend style={at={(0.5,1)},
                anchor=south,legend columns=3,legend cell align=left},
            symbolic x coords={Naive Bayes,Decision Tree,Random Forest},
            xtick=data,
             x tick label style={anchor=north},
            nodes near coords align={vertical},
            ymin=30,ymax=101,
            ylabel={Accuracy},
        ]
        \addplot table[x=interval,y=Dataset1]{\evr};
        \addplot table[x=interval,y=Dataset2]{\evr};
        \addplot table[x=interval,y=Dataset3]{\evr};
        \legend{Dataset 1, Dataset 2, Dataset 3}
    \end{axis}
\end{tikzpicture}
\caption{Overall evaluation of Naive Bayes, Decision Tree and Random Forest after unlearning.}
\label{figure11}
\end{figure}
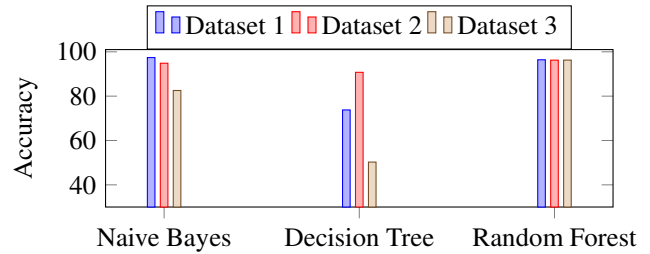

Figure \ref{figure11} demonstrates the restoring the accuracy by unlearning techniques. The polluted algorithms are unlearned, and thus, the accuracy of the unlearned algorithms become similar to the earlier. The accuracy of the Naive Bayes algorithm reaches to 97.39\%,	94.85\% and	82.55\% after unlearning for Dataset1, Dataset2, and Dataset3 respectively. Similarly, the accuracy of the Decision Tree algorithm reaches to 73.78\%,	90.75\%, and 50.27\% after unlearning for Dataset1, Dataset2, and Dataset3 respectively. However, the Decision Tree algorithm is unable to restore the original accuracy in unlearning process. The accuracy of the Random Forest algorithm reaches to 96.4\%,	96.24\%, and 96.26\% after unlearning for Dataset1, Dataset2, and Dataset3 respectively. It is observed that the unlearning accuracy of Random Forest slightly lower than the original accuracy. However, the difference between original and unlearning accuracy are significantly small.

\subsection{Overall performance}

\pgfplotstableread[row sep=\\,col sep=&]{
interval& 10 percent& 30 percent\\
Naive Bayes&	7.86&	2.17\\
Decision Tree&	8.56&	2.33\\
Random Forest&	9.17&	2.62\\
}\perf
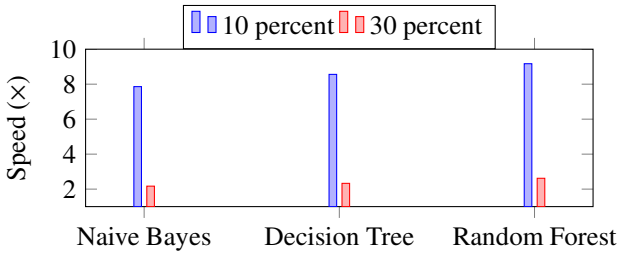
\begin{figure}[!ht]
\centering
\begin{tikzpicture}
    \begin{axis}[
          ybar,
            bar width=.1cm,
            width=0.45\textwidth,
            height=.2\textwidth,
            enlarge x limits=0.15,
            legend style={at={(0.5,1)},
                anchor=south,legend columns=3,legend cell align=left},
            symbolic x coords={Naive Bayes,Decision Tree,Random Forest},
            xtick=data,
             x tick label style={anchor=north},
            nodes near coords align={vertical},
            ymin=1,ymax=10,
            ylabel={Speed ($\times$)},
        ]
        \addplot table[x=interval,y=10 percent]{\perf};
        \addplot table[x=interval,y=30 percent]{\perf};
        \legend{10 percent, 30 percent}
    \end{axis}
\end{tikzpicture}
\caption{Overall performance in unlearning of Naive Bayes, Decision Tree and Random Forest over relearning.}
\label{figure12}
\end{figure}

Figure \ref{figure12} exhibits the overall performance of unlearning of the Naive Bayes, Decision Trees and Random Forest over retraining. Random Forest is the fastest among the three algorithms and Naive Bayes is the slowest among the three algorithms. Unlearning Random forest is $9.17\times$ and $2.62\times$ faster than the retraining the polluted machine learning algorithm in 10\% and 30\% pollution respectively. Similarly, unlearning Decision Tree is $8.56\times$ and $2.33\times$ faster than the retraining the polluted machine learning algorithm in 10\% and 30\% pollution respectively. Likewise, unlearning Naive Bayes is $7.86\times$ and $2.17\times$ faster than the retraining the polluted machine learning algorithm in 10\% and 30\% pollution respectively. It unveils that the machine learning algorithm spent much times in relearning in pollution. Therefore, unlearning the trained algorithms is always a better option than the retraining the machine learning algorithms in pollution as shown in Figure \ref{figure12}. Our experimental results shows that the unlearning is many times faster than the retraining the polluted algorithms.

\section{Discussion}
\label{Dis}
Restoring privacy, security, and usability is the goal of unlearning or forgetting systems as a whole. Even after unlearning polluted data, they aren't concerned with preserving any of the data that's already been entered into the system. The system's private data can still be exposed, the system's contaminated training data can still mislead prediction, and the system's faulty analytics can lead to erroneous suggestions. We showed how unlearning could be used in unlearning spam mail detection models, but to implement unlearning in a problem, we need to identify the problem associated with it; as each situation is idiosyncratic, researcher can come up with a model architecture that can unlearn data incrementally. Occasionally, the issue is simple to resolve. Some users may be aware of which of the user's data items are sensitive and would not wish to be reminded of those things in future interactions, for example. However, there are occasions when this becomes problematic. For example, a diligent operator can notice an unexpected decrease in spam detection rate, examine the event, and discover that some training data has been contaminated. We believe our approach to unlearning spam mail detection models are beneficial and can be integrated into the majority of email spam detection models to act instantly when polluted data is trained and unlearn the polluted data without retraining the entire model, which is paradigmatic, given the enormous size of the complete dataset and its inability to be retrained easily. An approach is presented for unlearning known polluted data, where the system owner knows which data is polluted and need unlearning. Still, in real-world scenarios, most of the time, it is unknown how models accuracy is harmed and which data must be removed to restore the model to its previous state, so, work can be done on identifying the pollution data due to which the model is being harmed and also to unlearn it automatically without human effort. Here, it is explained how unlearning could be performed on Naive Bayes, Decision trees, and Random forests. Researcher can develop architectures for every machine learning algorithm where unlearning is possible; for instance, unlearning on deep neural network where unlearning is done on neural networks. Only one model, in this case, has to unlearn the polluted data, but nowadays, once data is generated, it is trained to another model, and that data is trained to yet another model, and so on. However, in the approach presented here, only the first model forgets data; as part of future work, we invite other researchers to develop a solution to unlearn the polluted data across all models.

\section{Applications}
Machine unlearning is needed and can be used in various situations to preserve privacy, security and usability of a system. To preserve privacy, if a company changes it's terms and conditions regarding the use of user's data, the previous users might want their data to be forgotten by the machine learning systems which have been trained by the user's data from many days. In this case, systems must be able to forget some specific data which the user no longer wants it to be used by that particular company. In case of security, this work, unlearning in spam detection models comes under preserving security as many attackers try to exploit the model by polluting the data which is trained to the model. In this case, a model must be able to unlearn the polluted data without the need for retraining the healthy data. Not only in case of spam detection models, unlearning is needed in various situations which includes Malware detection models, trojan detection models and more.    When usability is considered, in case of Instagram, recommendation model tries to recommend posts based on previous search history. In some cases, the user might want to stop recommendations of a specific type of posts. So to solve this problem, model must forget the post which user previously liked, to recommend similar type of posts again to that user. Whereas, in the absence of unlearning, retraining can not solve the problem practically. As while retraining the model, for just forgetting small amount of data, all the previous data must be trained again which takes lots of time making it unfeasible to retrain. So unlearning serves a major function in this scenario.

\section{Conclusion}
\label{Con}
Our vision is discussed on how spam detection models can simply, quickly, and completely unlearn contaminated data, as well as the advantages of unlearning versus retraining polluted data. These methods can be tweaked a bit and used in a variety of unlearning situations. As it is demonstrated that not only in the case of security, unlearning can also be used to safeguard privacy and usability (in the case of spam detection models). In the instance of usability, imagine a movie recommendation system in which a user can label a movie as "Not interested," in which case the movie must be unlearned from the recommendations model so that it and similar movies are not suggested to the user again. In the event of privacy, the user can request that recommendation systems forget his data at any moment if the terms and circumstances change. The model is prepared to completely forget the specific user's data to preserve data privacy in such a situation.

\bibliography{mybibfile}

\end{document}